\documentclass[runningheads]{llncs}
\newcommand{\reffig}[1]{Fig. \ref{#1}}
\usepackage{graphicx}
\usepackage{subfigure}
\begin{document}
\title{Control of Pneumatic Artificial Muscles with SNN-based Cerebellar-like Model\thanks{First Author, Second Author and Third Author contribute equally to this work.\\}}
\titlerunning{SNN-based Control of Pneumatic Artificial Muscles}
%
\author{Hongbo Zhang\inst{1}\orcidID{0000-0002-0769-9243} \and
Yunshuang Li\inst{1}\orcidID{0000-0002-0941-4061} \and
Yipin Guo\inst{1}\orcidID{0000-0003-4335-2826} \and
Xinyi Chen\inst{2}\orcidID{0000-0002-3126-8953} \and
Qinyuan Ren\inst{3}\orcidID{0000-0001-9487-2675}}
%
%
\authorrunning{Hongbo Zhang, Yunshuang Li, Yipin Guo et al.}
\institute{College of Control Science and Engineering, Zhejiang University, 310013, China}
\maketitle              

\begin{abstract}
Soft robotics technologies have gained growing interest in recent years, which allows various applications from manufacturing to human-robot interaction. Pneumatic artificial muscle (PAM), a typical soft actuator, has been widely applied to soft robots. The compliance and resilience of soft actuators allow soft robots to behave compliant when interacting with unstructured environments, while the utilization of soft actuators also introduces nonlinearity and uncertainty. Inspired by Cerebellum’s vital functions in control of human’s physical movement, a neural network model of Cerebellum based on spiking neuron networks (SNNs) is designed. This model is used as a feed-forward controller in controlling a 1-DOF robot arm driven by PAMs. The simulation results show that this Cerebellar-based system achieves good performance and increases the system's response speed.    

\end{abstract}

\section{Introduction}
 Pneumatic artificial muscles (PAMs), such as Mckibben, are designed with the inspiration of creatures, showing great compatibility to creatures. This kind of muscle emerged in the twentieth century and has various kinds of applications after decades of development. Mckibben is small in size and relatively safe with high power to weight ratio. However, its nonlinearity and viscoelasticity properties increase the difficulty in controller designing. 
 
 The best example of control system for soft actuators can be found in animal bodies. As mentioned in \cite{Cerebellum}, Cerebellum as part of creatures' neural system have attracted vast attention because of that they play important role in controlling function. Therefore, we proposed a Cerebellum-like controller based on its real structure and internal information processing mechanism. Due to the bionic advantages of this controller, it's appropriate to apply it in controlling Mckibben. 
 
 In our system, we conduct a Cerebellum-like controller based on spiking neural networks (SNNs) to control a 1-DOF robotic arm shown in \reffig{robot_arm} driven by a pair of PAMs. It can also be refered as an online closed-loop error-correction controller. The controller has one kind of fibers, Mossy Fibers, with four kinds of cells, Granule cells, Purkinje cells Inferior Olive cells and Deep cerebellar nuclei cells. All of them are constructed as the similar structures of Cerebellum and we model its physical functions. Besides, the SNNs functions the feed-forward part in our controller. Previously, scholars have designed some novel SNNs\cite{SNNModel1}\cite{SNNModel2}\cite{SNNModel3}.In our work, a new SNN topology is designed to learn the inverse model of soft actuators and make up for the output of the controller.


\section{The Structure of SNN}
\begin{figure}[htbp]
\centering
\begin{minipage}[t]{0.44\textwidth}
\centering
\includegraphics[width=5.4cm]{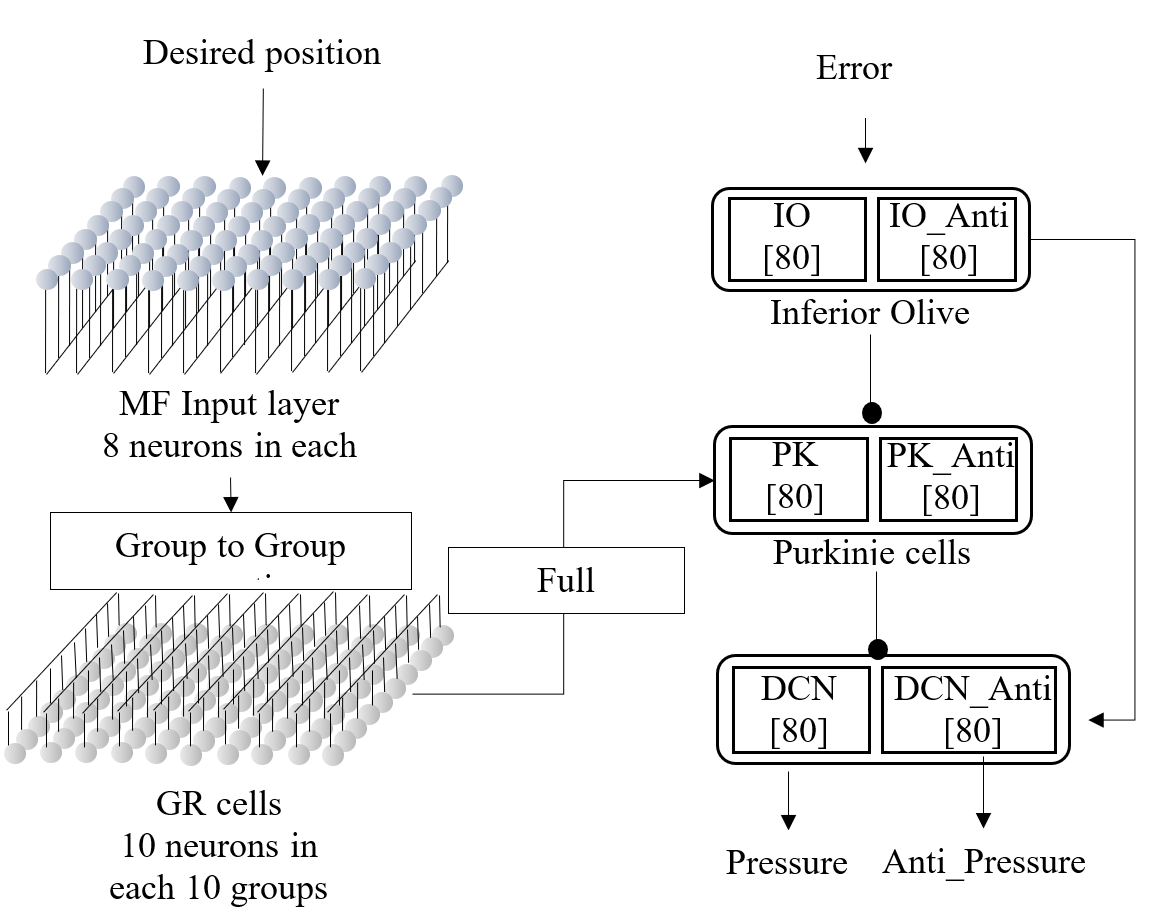}
\caption{Topological structure diagram of the neural network. The arrow indicates an excitatory effect, and the circle indicates an inhibitory effect.}
\label{topology}
\end{minipage}
\begin{minipage}[t]{0.44\textwidth}
\centering
\includegraphics[width=3cm]{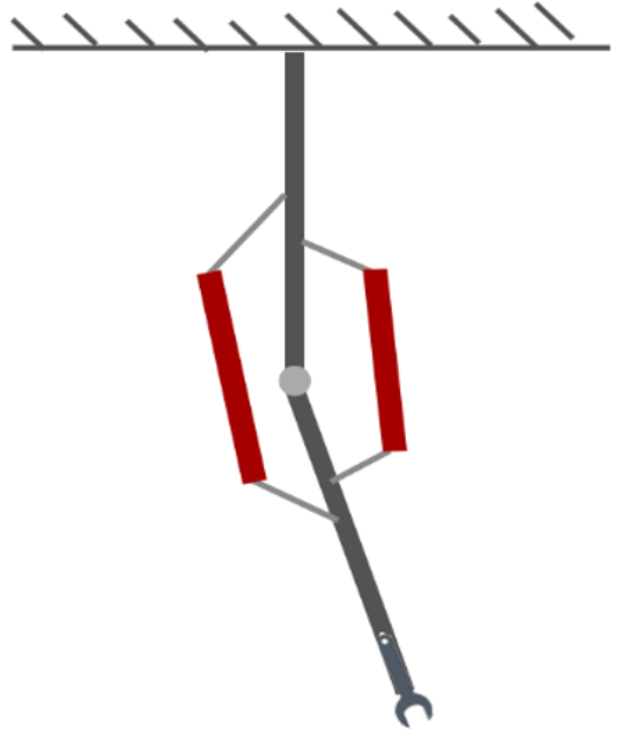}
\caption{The robotic arm. Two pneumatic artificial muscles are placed on both sides. When one of them contracts, it pulls the robotic arm to rotate.}
\label{robot_arm}
\end{minipage}
\end{figure}

Here we use a real-time spiking neural network with a cerebellar-like structure that can obtain the inverse model of the Mckibben pairs to act as a feed-forward part of the controller. We use a set of spiking neurons as a basic unit of the network and imitate the structure of Cerebellar to build a neural network, which bases on \cite{SNNModel1}. The topology of the network is displayed in \reffig{topology}.

\subsection{Neurons and Layers}
The neural network consists of about 1000 neurons  which contains 80 Mossy Fibers (MF), 100 Granule cells (GR), 160 Purkinje cells (PK), 160 Inferior Olive cells (IO), and 160 Deep cerebellar nuclei cells (DCN). The leaky integrate-and-fire (LIF) neuron model is used to build the neurons.
Since the cerebellar cortex has hierarchical functional blocks, different blocks are responsible for different types of physical movements. The GR layer is connected to the MF layer hierarchically to imitate this partition mapping pattern. The PK layer and the DCN layer are divided into antagonistic pairs to receive pulses from GR and corresponding error signals from IO. Weights between GR and PK are adjusted and trained according to Spike Timing–Dependent Plasticity (STDP) learning rules. 

\subsection{Learning Rules}
Studies have shown that learning in Cerebellum is mediated by synaptic plasticity. In our network, the connection between the GR layer and the PK layer is carried out according to the learning rules.

STDP is a learning method based on Hebbian learning rules. The dependence of synaptic modification on the order of pre and postsynaptic spiking within a critical window of tens of milliseconds has profound functional utilities in learning and memory.\cite{Caporale2008SpikeTP}

STDP in this network is divided into long-term potentiation (LTP) mediated by GR pulses and long-term depression (LTD) mediated by IO pulses. The equations are as follows:
LTP effect increases the weight $w$ at a specific learning rate whenever there is a GR pulse, while the LTD effect adds the historical GR pulse to the kernel function whenever there is an IO input pulse. 
\begin{equation}
{\rm LTP}:{\rm \Delta} w\left( t \right) = {\rm nu_ {0}~}\delta\left( t \right),
\end{equation}
\begin{small}
\begin{equation}
{\rm LTD}:{\rm \Delta} w\left( t_{\rm IO} \right) = ~ - {\rm nu_{1}}{ \int_{\rm - \infty}^{t_{\rm IO}}{K\left( {t - t_{\rm IO}} \right){\rm\delta}_{\rm GR}\left( t \right){\rm dt}}},
\end{equation}
\end{small}
\begin{equation}
K(x) = {\rm e}^x-{\rm e}^{4x}.    
\end{equation}
where ${\rm nu_{1}, nu_{2}}$ represents learning rate for LTP and LTD respectively, $\delta$ represents pulse signal of the corresponding neurons and $K$ represents the kernel function described in Eq.(3).

\section{Simulation Platform}
Bindsnet\cite{bindsnet}, an open-source spiking neural network building platform, is used to build and train the neural network. Meanwhile, Simulink is used as the platform to build the physical simulation environment and the controller.

\subsection{The Robotic Arm}
Our robotic arm uses two iron rods as bones and a pair of pneumatic muscles as actuators imitating bicep and tricep of human respectively in \reffig{robot_arm}.

One end of the link is fixed, and the muscles are installed on both sides of the link. When one of the muscles contracts, it will pull the unfixed link to rotate. We take the deflection angle of the robotic arm as output and model the robotic arm in Simulink.


\subsection{Control Loop}
\begin{figure}
\centerline{\includegraphics[width=3.5 in]{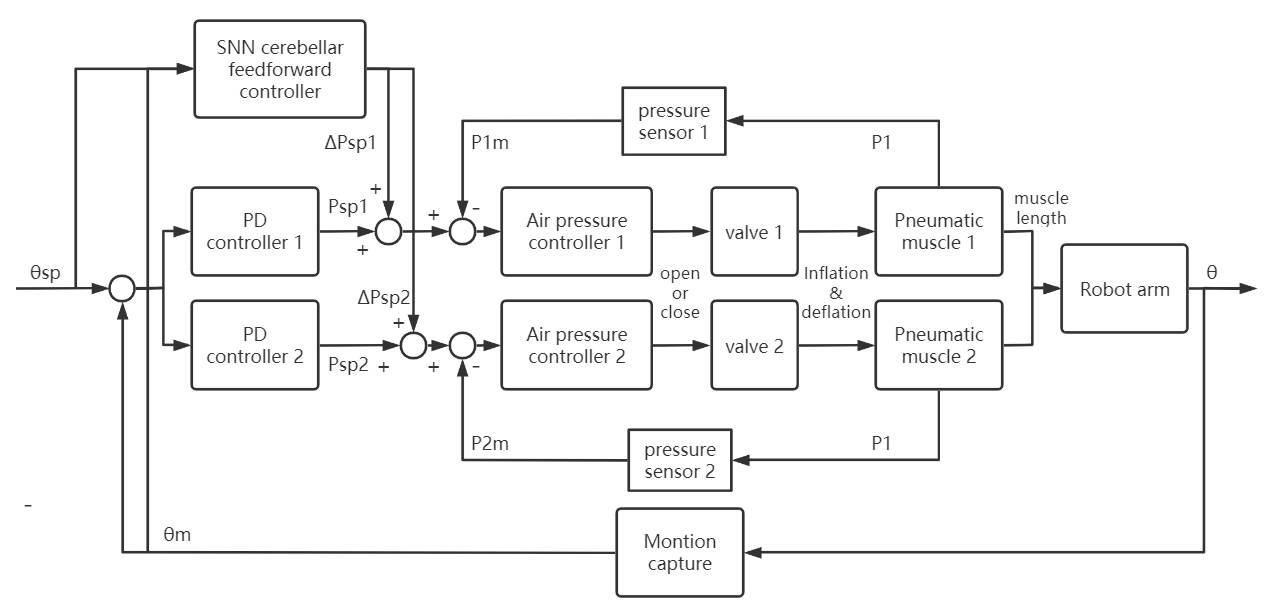}}
\caption{The block diagram of the feed-forward and feedback cascade control system.}
\label{control_system}
\end{figure}

A cascade control method including a feed-forward part and a feedback part is applied in our system shown in \reffig{control_system}
\vspace{-0.5cm}
The air pressure feedback controller shown in \reffig{control_system} directly controls the air pressure of the two pneumatic muscles by controlling opening and closing of the solenoid valves.

The PD controllers in the outer loop is served as the main controller to obtain the precision of motion control. Cerebellum-inspired feed-forward controller contributes to improve the response speed and deal with the nonlinearity.


\section{Results and Discussions}
\begin{figure}[htbp]
\centering
\label{Results}
\subfigure[]{
\begin{minipage}[b]{0.2\linewidth}

\centerline{\includegraphics[width=4.5 in]{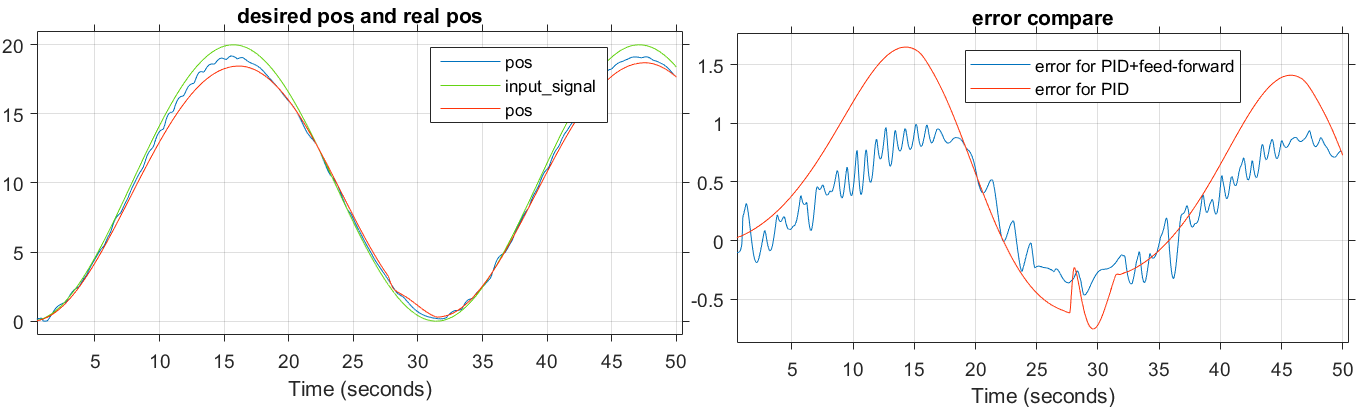}}
\end{minipage}%
}%

\subfigure[]{
\begin{minipage}[b]{0.2\linewidth}
\centerline{\includegraphics[width=4.5in]{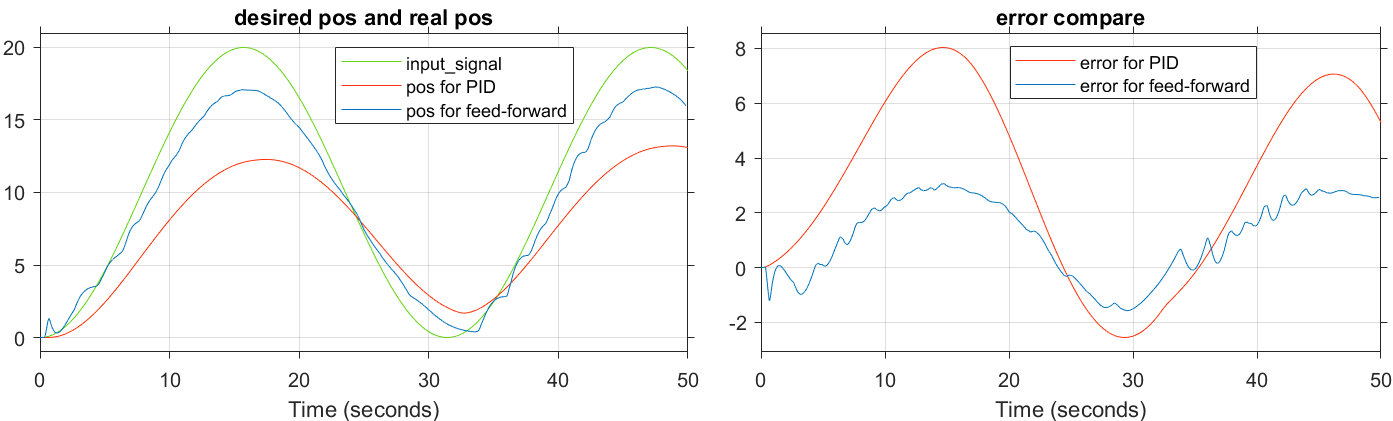}}
\end{minipage}%
}%
\caption{(a) Results for a controller with both feed-forward and feedback blocks. (b) Results for a controller with a single feed-forward block.}
\end{figure}

In the simulation, we use two different control strategies:(1) A controller with both feed-forward and feedback blocks (2) A controller with a single feed-forward block in \reffig{Results}.

Firstly, a controller with a PD feedback part and a feed-forward part built by the network is applied. The feedback controller is added to achieve the rapid response of the control system to disturbances. In order to test the trajectory tracking effect and the anti-interference effect of the end of the manipulator, a sinusoidal trajectory input is applied as the desired trajectory to analyze the control effect.Results in \reffig{Results} (a) shows improvement in control accuracy comparing with the PD controller. Results in \reffig{Results} (b) also indicate that the feed-forward controller achieves good performance as well.

The experiment verifies that our controller can replace traditional controllers, and we will continue to reduce the effect of feedback part and verify the control effect of our controller on Mckibben artificial muscle. The whole system shows strong bionics and has potentially large applications in many fields.

\bibliography{reference.bib}

\begin{thebibliography}{1}
\providecommand{\url}[1]{\texttt{#1}}
\providecommand{\urlprefix}{URL }
\providecommand{\doi}[1]{https://doi.org/#1}

\bibitem{SNNModel2}
Abadía, I., Naveros, F., Garrido, J.A., Ros, E., Luque, N.R.: On robot
  compliance: A cerebellar control approach. IEEE Transactions on Cybernetics
  \textbf{51}(5),  2476--2489 (2021). \doi{10.1109/TCYB.2019.2945498}

\bibitem{Caporale2008SpikeTP}
Caporale, N., Dan, Y.: Spike timing-dependent plasticity: a hebbian learning
  rule. Annual review of neuroscience  \textbf{31},  25--46 (2008)

\bibitem{SNNModel1}
Carrillo, R.R., Ros, E., Boucheny, C., Coenen, O.J.M.: A real-time spiking
  cerebellum model for learning robot control. Biosystems  \textbf{94}(1),
  18--27 (2008). \doi{https://doi.org/10.1016/j.biosystems.2008.05.008},
  \url{https://www.sciencedirect.com/science/article/pii/S0303264708001226},
  seventh International Workshop on Information Processing in Cells and Tissues

\bibitem{bindsnet}
Hazan, H., Saunders, D.J., Khan, H., Patel, D., Sanghavi, D.T., Siegelmann,
  H.T., Kozma, R.: Bindsnet: A machine learning-oriented spiking neural
  networks library in python. Frontiers in Neuroinformatics  \textbf{12}, ~89
  (2018). \doi{10.3389/fninf.2018.00089},
  \url{https://www.frontiersin.org/article/10.3389/fninf.2018.00089}

\bibitem{Cerebellum}
Miall, R.: The cerebellum and visually controlled movements. In: IEE Workshop
  on Self-Learning Robots III Brainstyle Robotics: The Cerebellum Beyond
  Function Approximation (Ref. No. 1999/049). pp. 2/1--2/5 (1999).
  \doi{10.1049/ic:19990257}

\bibitem{SNNModel3}
Yang, J., Song, T.: A prediction scheme in spiking neural network (snn)
  hardware for ultra-low power consumption. In: 2020 International SoC Design
  Conference (ISOCC). pp. 310--311 (2020).
  \doi{10.1109/ISOCC50952.2020.9333106}

\end{thebibliography}

\end{document}